# Towards an architectural framework for intelligent virtual agents using probabilistic programming

**Anton Andreev[1], Grégoire Cattan[2]**

[1]  Université Grenoble Alpes, CNRS, Grenoble INP, GIPSA-lab, 38000 Grenoble, France; andreev.anton@gipsa-lab.grenoble-inp.fr
[2]  IBM Software, Data & AI, Kraków, Poland; gregoire.cattan@ibm.com

**Abstract:** We present a new framework called KorraAI for conceiving and building embodied conversational agents (ECAs). Our framework models ECAs' behavior considering contextual information, for example, about environment and interaction time, and uncertain information provided by the human interaction partner. Moreover, agents built with KorraAI can show proactive behavior, as they can initiate interactions with human partners. For these purposes, KorraAI exploits probabilistic programming. Probabilistic models in KorraAI are used to model its behavior and interactions with the user. They enable adaptation to the user's preferences and a certain degree of indeterminism in the ECAs to achieve more natural behavior. Human-like internal states, such as moods, preferences, and emotions (e.g., surprise), can be modeled in KorraAI with distributions and Bayesian networks. These models can evolve over time, even without interaction with the user. ECA models are implemented as plugins and share a common interface. This enables ECA designers to focus more on the character they are modeling and less on the technical details, as well as to store and exchange ECA models. Several applications of KorraAI ECAs are possible, such as virtual sales agents, customer service agents, virtual companions, entertainers, or tutors.

**Keywords:** embodied conversational agent; intelligent virtual assistant; HCI; virtual agent; virtual assistant; virtual actor; virtual human; probabilistic models; probabilistic programming; probabilistic programming language

## 1. Introduction

An embodied conversational agent (ECA) is an animated computer-generated character capable of performing human-like communication with human partners and showing autonomous behaviors (1). ECAs emulate different human cognitive abilities, such as speech, listening, grabbing attention, displaying emotions, and so forth. These agents can take on several roles, for example, as an assistant, a tutor, an information provider, or a customer service agent. As noted in (2), the success of these virtual agents depends on the level of human-like interaction they provide. Examples of ECAs are available in (1, 3, 4), and extensive surveys can be found in (2,5). ECAs can also be used in the medical domain for mild sleep disorders and addictions (6) and to help elderly people with cognitive impairments (7). In the future, ECAs may even be used as close friends or (intimate) partners. This use is often depicted in science fiction movies. Examples are the ECA Joi in *Blade Runner 2049* (Denis Villeneuve, 2017) and Cortana in the *Halo* games franchise initiated by Bungie, Inc.

In this paper, we view human behavior from the perspective of inner physical needs, that is, hunger, thirst, sleep, and social interaction in the form of, for example, expressing an emotion or opinion, providing advice, asking for information or advice, telling a joke, and so forth. Behavior can be viewed as an internal distribution of needs and desires interconnected with more complex (probabilistic) models that are continually updated by the environment. Interactions with the environment should not be seen as the only source of updates. For example, a person can change its behavior to bored or tired while waiting in a queue. There was no interaction, but the fact that time has passed has affected the person's behavior. In (8), the authors assumed that human learning and inference approximately follow the principles of Bayesian probabilistic inference and explore the conceptual and mathematical foundations of Bayesian models of cognition. KorraAI also focuses on distributions and Bayesian networks and how they are updated for the implementation of cognitive processes.

KorraAI is intended for domains such as e-commerce, entertainment, education, and e-health. The conceived agent can be a virtual babysitter, a companion, a personalized media content presenter, an electronic caregiver (ECG), a professional coach, or a character in video games.

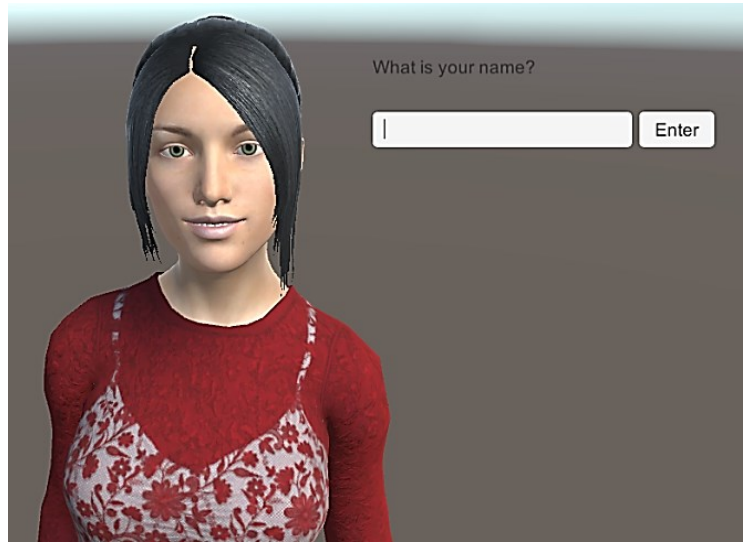

**Figure 1**. Representation of Joi, a female ECA implemented using KorraAI.

In this paper, we also present a concrete implementation of the framework. We have implemented a female ECA called Joi (**Figure 1**), modeled to be an everyday companion and entertainer. Joi can ask questions, tell jokes, and provide advice and recommendations. It is designed to be present in the user's everyday life. For example, humans can interact with Joi before work and after or during the day if they stay at home. Joi could also be used to keep an elderly or lonely person company. In the remainder of this paper, we present the advantages of the KorraAI framework, using Joi as a case study. The source code of the Joi model can be obtained on demand.

*1.1. Existing ECA frameworks*

As explained in (9), ECA systems are difficult to develop for individual research groups. Creating an ECA requires the integration of many competences and components that must be carefully coordinated in order to achieve coherent human-like behavior. These include input methods such as automatic speech recognition (ASR), gesture recognition, natural language processing (NLP), information extraction (IE), intent detection, efficient encoding of knowledge, models of concepts in a specific domain or life in general, objectives, reasoning, and visual 3D embodiment that supports facial expressions, microexpressions, hand gestures, mimicry, and backchannels. The other, typically human communication competences that need to be implemented in an ECA are turn-taking and interlocutor interruption during a conversation. Different ECA architectures excel in specific areas and lack in others. At least three different ECA typologies can be distinguished:

- Based on the increased complexity of conversational styles. In (10), ECAs are classified into the following four conversational styles: "TV Style Presenters", "Virtual Dialogue Partners", "Role Plays and Simulated Conversations", and "Multithreaded Multiparty Conversation".
- Generic or Specialized. Specialized ECAs are used in specific application domains. They are easier to create because linguistic and nonverbal vocabularies are focused on one domain or scenario (e.g., online sales), making, for example, speech recognition and synthesis models easier to tune. Generic ECAs can be used in several domains, so they need to be able to handle various topics in conversations and nonverbal interactions.
- Partial or Full. Partial ECAs implement only certain functionalities, for example, speech recognition and synthesis. Indeed, a virtual partial agent that provides customer service might not need to display emotions to assist clients. This is the difference compared to a full ECAs, which tries to implement the full spectrum of human cognitive abilities.

The potential of ECAs has been explored in the e-health domain. For example, in (11), a virtual agent called Louise, implemented as a smart-phone application named KANOPEE, was used to manage sleep disorders. Such mobile virtual companions can be very helpful in treating the lighter phases of a medical condition. They can provide fast access to health assistance while constantly evaluating the user's state and, if needed, recommend contacting a human expert in the domain. A type of ECAs called relational agents (RA) are described in (12). These RAs have the objective of establishing long-term socio-emotional relations and, in particular, trust, rapport and therapeutic alliance with the user in order to improve adherence to treatment. In the case of mental illness, the authors highlight that the social behavior of these agents may provide many of the benefits of therapeutic alliance and social support these patients need without the anxiety, stigma, or demands of face-to-face conversation with other people. Potentially, ECAs could also be used as psychotherapists or to boost confidence. Several existing ECA architectures, for example, MAX (13), use the belief-desire-intention (BDI) models (14). "Belief" represents the agent's model of the world, "desire" represents the agent's goal(s), and "intention" denotes the action choice. The framework "How Was Your Day?" (HWYD) (15) is a companion ECA with whom the user discusses how his day has passed. The discussion takes the form of free conversation on a limited number of topics. The ECA provides advice and support to the user, considering the emotions expressed by the user through dialogue. In (9), a general ECA framework called GECA is proposed. It allows for communication between different ECA components written in different programming languages. It also provides an interaction language similar to Artificial Intelligence Markup Language (AIML) (16) with several improvements. Similar component integration projects are UECA (17) and whiteboards (18). Another example is Greta (3,19), a virtual agent capable of expressing verbal and nonverbal behavior, as well as listening, that uses the SAIBA framework (20). SAIBA presents a general representational framework for real-time multimodal behavior generation at the macroscale. According to SAIBA, an ECA should be composed of three modules—an intention planner, a behavior planner, and a behavior realizer—that communicate with each other using two XML-based languages: Functional Markup Language (FML) (21) and Behavior Markup Language (BML) (20). Several ECA realizations provide their own implementations of this framework. For example, Greta uses the FML-APML (21) to specify the agent's communicative intentions (e.g., its beliefs, emotions, etc.). Another BML realizer called AsapRealizer is presented in (22). ECAs can also be used for teaching, as in (23), or as virtual patients for medical students, as in (24). The Virtual Human Toolkit (VHToolkit) (25) is a framework comprising five modules: a multimodal sensing framework, a module for selecting the character's responses, a nonverbal behavior generator (NBG), SmartBody (25) animation library, and Unity 3D integration. Similarly to Greta (3) and LOUISE (7), it uses BML to handle nonverbal behavior. Agents United (26) provides a common platform for researchers and developers to set up their own ECAs and allows the ECAs to seamlessly interact with each other. It is built on an architecture of four layers: sense, remember, think, and act. It also includes a built-in interaction manager, Flipper (27); two dialog execution systems; and two multi-model behavior realizers, ASAP (22) and GRETA.

Currently, a promising technology for developing ECAs is the open-source Babylon.js[1] 3D framework. Some open-source projects, such as Amazon Sumerian Hosts,[2] allow for easy integration with Babylon.js and Amazon's AI services, such as Polly (speech-to-text generation) and Amazon's Lex that is used to build conversational bots using Amazon's automatic speech recognition ASR and natural language understanding (NLU) services. Using commercial cognitive services from companies like Amazon, Google, and Microsoft can help build an ECA much faster with, for example, a quality text-to-speech, but it poses a risk when a company's policy shifts and a product or service is deprecated. Moreover, these services depend on an internet connection and are generally paid.

### 1.2. Limitations and challenges of currently available ECAs

An ECA could be based on a conversational agent that uses decision trees or could be rule-based. Decision trees are often used by customer support agents, where the user is guided through questions to the solution to his problem. Rule-based chatbots are provided with a database of responses and a set of rules that help them match an appropriate response from the provided database. Such restricted rule sets frequently limit the model's capability and can result in low-accuracy answers and an unpleasant user experience, as highlighted in (28). Every time the user is subjected to the same communication paths, this may decrease the agent's believability and engagement. Therefore, an open question is, "How does one add variability to the communication?" There are some human norms for interaction. For example, one should not always talk about the same subject, change the topic of the conversation, and, as a whole, perform interactions that the human partner will not get easily accustomed to or that will at least not feel unnatural. Variability can also be added to the construction of the phrases used to communicate with the user. The objective is to improve both believability and engagement.

---

[1] https://www.babylonjs.com
[2] https://github.com/aws-samples/amazon-sumerian-hosts

Another subject of research is how the ECA can initiate commutation. This is the ability to express proactive behavior. In practice, many ECAs react only to humans, who always start the interaction. ECAs are able to respond, but they are unable to initiate interactions by themselves. Furthermore, the lack of an answer to a question posed by an ECA should not block it forever, but it should show an interaction initiative and continue with another utterance after a certain timeout.

Another challenge is modifying behavior. An ECA should be able to, for example, increase the number of suggestions it provides and decrease the number of questions it asks to get to know the user in a way that adjusts all other encoded behaviors accordingly. Such modifications should be adjustable over time.

Depending on the design, an ECA will try to learn the user's preferences in order to be more useful and/or likable. We postulate, however, that an ECA's behavior should not be completely dependent on the user's preferences; there should also be some probability of showing opposite or unexpected behaviors in order to provoke surprise. This, by itself, can contribute to a more natural ECA.

*1.3. Usage of Probabilistic Models and Probabilistic Programming*

Usually, in computer programming, variables are assigned specific values (e.g., an "age" variable). Probabilistic programming (PP), on the other hand, allows one to represent random variables; their values are not known precisely, so they are represented by a probability distribution over the value rather than by a single fixed value. Distributions are used to capture uncertainty. For example, when people meet a new person, they are more likely to assign a range over her estimated age instead of an exact number. KorraAI is an ECA framework that uses probabilistic models and inference in the form of well-studied statistical distributions or more complex Bayesian models. Probabilistic models (PM) can be non-linear and can be compactly written as code using PP. At the same time, modeling and inference are separated, which makes them easier to employ. Programs developed using PP are considered as non-deterministic because even if the inputs are the same, the outputs might differ. A general method of transforming arbitrary programming languages into PPL with a straightforward MCMC inference engine is described in (29). In (30) the authors experiment with three PPLs (Anglican, WebPPL and Figaro) for the recognition of casual, structured human activities. They praise the fact that PPLs are very convenient for modeling casual models. They point that one needs to be aware of the different capabilities of each probabilistic inference algorithm and that for certain tasks, domain specific inference algorithms can outperform the generic ones provided by most PPLs. This can be improved by adding more domain specific inference algorithms to PPLs. The authors state that using PPLs trades off a (potentially) increased inference time for a faster model development.

In some frameworks, only certain ECA functionalities are implemented using PM. For example, it can be an ASR task (31), an NLP task of probabilistic part-of-speech tagging (32), or a multi-modal interlocutor-aware generation of facial gestures (33) based on normalizing flows. Normalizing flows provide a general way of constructing flexible probability distributions over continuous random variables. Probabilistic models are also used in (34) to predict a listener's backchannels using the speaker's multimodal output features (e.g., prosody, spoken words, and eye gaze). This is similar to (19), where probabilistic rules are used to estimate the best backchannel response to be performed by the agent. PP can be used in the domain of robotics, as in (35). The authors propose the ROSPPL framework that could perform automatic state estimation, fault detection, and parameter calibration. It uses a PPL called WebPPL[3] (36). The advantage of this approach lies in its generality, which makes it useful for quickly designing and prototyping new robots.

PM can also be used for higher-level cognitive tasks. Some machine learning algorithms require high volume of examples to train. Humans, on the other hand, require a much smaller number of examples; sometimes, just a few examples are sufficient. PM (Bayesian learning) allows for the specification of prior knowledge precisely, therefore requiring fewer data to learn. Moreover, PM can be updated over time. Priors can be set by an ECA designer, and the models can be updated during a phase of adaptation toward the user. Behavior was modeled in (37) for the control of believable characters in video games. The authors propose an improved version of the input-output hidden Markov model (IOHMM) for imitation learning, which was first presented in (38). In imitation learning, the agent learns its behavior using observations of one or several players. In the IOHMM behavior model, hidden states are decisions, inputs are stimuli, and outputs are actions. Mood estimation is modeled with PM in (39) using dynamic Bayesian networks (DBNs). The book (40) demonstrates practical examples of behavior modeling using WebPPL. The authors state that while knowledge often changes gradually as data is accumulated, it sometimes jumps in non-linear ways. An example of a probabilistic model that captures that non-linearity is given in section "Learning as Conditional Inference". It models a posterior distribution over how likely the coin we are tossing is fair based on observed data and prior knowledge regarding the fairness of the coin. This enables the examination of several scenarios. For example, if a coin lands five times heads, it is more likely to be considered fair. Five heads can be a minor coincidence. If the coin lands 15 times heads, and even without any indication of tampering, the odds would strongly indicate the presence of a biased coin.

In the book[4] (41), WebPPL is again used to create a richer PM of human planning, which captures human biases and bounded rationality. The authors of (42) use PP for affective computing, modeling psychological-grounded theories as generative models of emotion and implementing them as stochastic, executable computer programs. The authors state that a probabilistic programming framework can provide a standardized platform for theory-building and experimentation. In (43), the authors use PP and models of the theory of mind to create chaser-runner behavior models, wherein a "chaser" pursues a "runner" on a predefined city map. These models study how to give autonomous agents the ability to infer the mental state of other agents and how to reason about this state for decision-making and planning. By reasoning about one's opponent, one can craft counterplans against them. The authors conclude that PP is a natural way to describe these models.

KorraAI uses PM and PP to, for example, process input from the user and, on a higher level, emulate human behavior and plan future interaction steps.

## 2. Core concepts and motivation behind KorraAI

One of the main ideas behind our framework is to switch from a hard-coded interaction pipeline, and event-based reactive interactions, to one that uses an ensemble of distributions to define the behavior of an ECA.

An "interaction" in KorraAI is an action to be performed by the ECA, and it can be asking a question, reacting, or making a statement, or it can be non-verbal, such as a change in appearance. A "category" is a group of similar interactions. The interactions are distributed over time. The question "What should the agent do next?" is addressed in our framework by picking an interaction to be executed by the ECA.

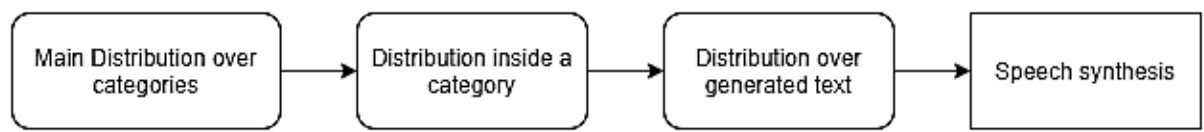

**Figure 2.** The behaviors produced by the KorraAI ECA depend on several distributions.

In **Figure 2**, we show how several distributions are used together, starting from selecting an interaction to the exact words used by an ECA. By using a range of distributions, KorraAI strives to produce a non-repetitive, variable, and hard-to-anticipate human-like sequence of interactions. KorraAI implements specific research and design objectives, as shown in **Table 1**.

---

**Table 1.** Summary of research objectives set in KorraAI and their implementation.

| Research or design objective | Implementation |
| --- | --- |
| Behavior that evolves over time | In KorraAI, distributions can be set to change after a specific elapsed time, not just based on the interaction with the user. See **Figure 3**. |
| Avoid fixed events | Distributions are used—from selecting the next interaction to calculating the delay between two interactions. Predefined sequences and fixed time intervals are avoided. Speech generation also employs distributions. |
| ECA is proactive and non-blocking | The ECA can start communication by itself and switch to another "interaction" if no response from the user is received. |
| An ECA running over a long period of time | A distribution can be set to take into account the time of the day and the user's daily schedule. KorraAI provides a tool assisting the creation of a distribution over a period of time. |
| Manage uncertainty in responses from the user | Predefined user responses are mapped to probabilities and used in the probabilistic distributions and models. |
| Explore the use of PPL | Advances in PPL allow for easier creation and use of probabilistic models. |
| Focus on the ECA model | An ECA is created using a plugin architecture that hides the technical details of the execution engine. KorraAI also provides a programming interface that facilitates the use of distributions and probabilistic models. |
| A dual model of ECA and human interaction partner | Bayesian networks include probabilistic variables and models that describe both the ECA and the user. The model of the user is used for adaptation. |
| Real-time execution | The system is able to respond uninterruptedly even when a heavy probabilistic reference is involved. |

To handle the objectives in **Table 1,** we used an architecture that can encode uncertainty in the user's response, which is then linked to distributions and Bayesian networks.

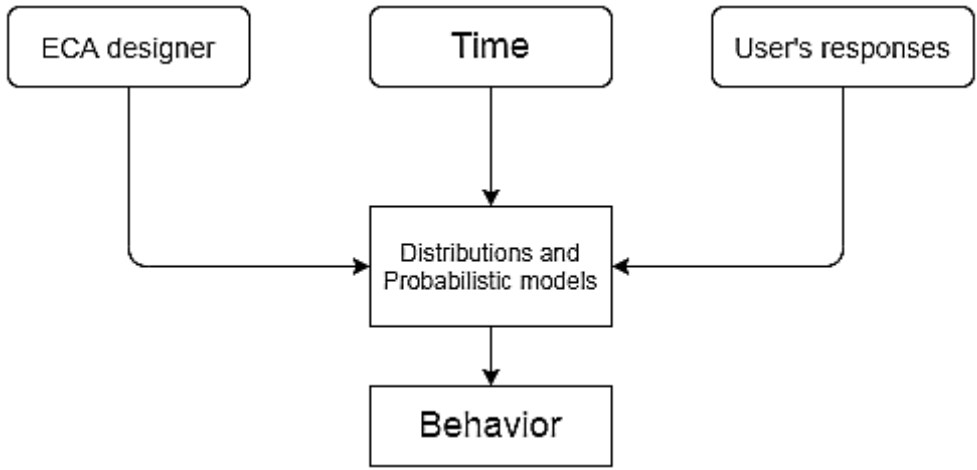

**Figure 3.** Simplified representation of KorraAI's architecture.

**Figure 3** provides a basic view of KorraAI's architecture. KorraAI targets the modeling of behavior patterns that depend on three factors: a predefined character, time, and the previous interaction with the user. The ECA's "interactions" are generated using distributions and evolve (or not) over time in order to simulate natural behavioral patterns. We call it "controlled indeterminism" when the consecutive interactions are not hard-coded but result from applying distributions. As shown in Section 3.7, some level of determinism in the ECA's behavior is still needed. The job of the ECA designer in KorraAI is to provide interactions and distributions and to model the internal states of the agent (e.g., emotions) using Bayesian networks. In KorraAI, the primary direction of communication is from the ECA toward the user.

A good application example of KorraAI is a virtual sales agent. The goal of a sales agent is to guide the conversation to reach a sales objective by performing a series of interactions. Such an agent can be implemented using several categories of interactions, such as "getting to know questions", "small talk", and "advertise a product". Then, a distribution is constructed over these three categories so that the ECA can start with a mix of the first two and gradually increase the third one. Another example is the construction of a virtual teacher. Its main objective is to teach the user a certain skill, but it must also maintain the user's attention. It can be modeled with three categories: "explanations", "questions regarding what the user has learned" and "small talk". Carefully defining a distribution over these categories is important from a cognitive perspective. For example, when the ECA executes a joke interaction as part of the "small talk" category, it allows the user to have some rest. Asking questions (the second category) during the lesson (the first category) can be considered as more gainful and engaging as opposed to asking all questions at the end of the lesson. It is the job of the ECA designer to provide a probabilistic distribution over the categories.

KorraAI is a Partial ECA. Per the classification given in (10), KorraAI falls between the categories of a TV-style presenter and a virtual dialog partner. It could be used as both a specialized ECA and a generic one.

### 2.1. Model

A model represents a single ECA. It has several components and defines discussion topics, reactions to user responses, and the ECA's behavior. Models can be stored separately from the execution engine, so at each start, a different model can be loaded. This process is explained in Appendix C.

### 2.2. Interactions

The behaviors in KorraAI are defined through groups of interactions, which are usually in the form of a brief dialog with the user. Once an interaction is executed, it is marked as "used" and "answered" if it was a question that received a response from the user. By default, it is not used again. An example of such an interaction is a question about the user's age.

### 2.3. Categories

Categories are groups of interactions. They can be considered as different topics of discussion, but they may also represent nonverbal behaviors. Categories are to be provided by the ECA designer. The following are some example categories:

*Initial Communication Introduce ECA.* In this category, all interactions are used to present the ECA: name, age, and some other personal details.

*Initial Communication Ask User*. Using the interactions in this category, the ECA will ask the user for some personal information (name, age, etc.). The questions can be either predefined or constructed using information already known by the ECA.

*Ask Uncertain Question*. This category contains questions used to collect some uncertain information. Usually, these are related to the state of the user and his preferences. Examples are "Are you tired?" "Do you like jokes?" and "Are you doing OK?" These questions can be used multiple times, following a distribution over time. This is because the state of the user and his preferences can change over time.

*Make a Suggestion*. This is a category that contains recommendations the ECA gives to the user. It can have several subcategories, for instance, "Propose movies to watch," "Health advice," and "Propose music to listen to."

*Make a joke*. Jokes can be a single statement or longer based on turn-taking. Examples include the ECA's question, the human response, and the ECA's final reaction. This three-turn action sequence is often used in KorraAI.

*Change Visual Appearance*. This category comprises several outfits that follow the distribution described in Section 3.6. In the ECA Joi, the change is announced verbally before execution. There is another distribution over the phrases in the announcement so that it is not the same at every outfit change. The outfit change can be considered a preference manifested by the ECA.

*Express ECA State.* This category is used to communicate the ECA's current internal state. It can be set to a physiological factor, such as "Sleepy" or "Tired," or more emotional, such as "In a good mood," "In an optimistic mood," "Sad," and so forth.

### 2.4. Main Distribution

The Main Distribution governs the overall behavior of the ECA. It is the top distribution that is used to decide which category should be used next.

| **Log Listing 1** |
| --- |
| **************** BEGIN Regenerating interactions **************** |
| Histogram: |
| MakeSuggestion 37.5% #################################### |
| AskUncertainFactQuestion 0.791% |
| AskPureFactQuestionAboutUser 27.7% ########################## |
| SharePureFactInfoAboutBot 31.6% ############################### |
| ChangeVisualAppearance 1.19% # |
| ExpressMentalState 1.19% # |

**Log Listing 1** is taken from KorraAI's log file and shows the result of probabilistic sampling of the Main Distribution over seven categories, where "AskPureFactQuestionAboutUser" is the equivalent of "Initial Communication Ask User" and "SharePureFactInfoAboutBot" corresponds to "Initial Communication Introduce ECA".

### 2.5. Interactions Queue

Interactions are periodically sampled and buffered in the Interactions Queue. **Log Listing 2** shows an automatically generated example of an Interactions Queue. The queue can be modified or regenerated during the interaction with the user.

| **Log Listing 2** |
| --- |
| Interactions queue: \|1. Hi \|2. What is your name? \|3. My name is Joi. \|4. I am 30 years old. \|5. How about this TV series Shameless? I recommend it. \|6. Time for some sport. You should go to the gym. Sports are good for both physical and mental health. \|7. <prosody pitch="+5%">I was thinking. God must love stupid people. <break time="600ms"/>He created SO many of them!</prosody> \|8. ###place holder for InAGoodMood \|9. How old are you?\| |

Some interactions are complete with text already pre-generated, but for others, such as "###place holder for InAGoodMood," only the category is available, and the exact text is filled later before actual execution depending on the state of the ECA.

### 2.6. Distributions

There are two types of distributions in KorraAI:

**1.** High-level distributions govern decision-making or higher-order cognitive processes. Examples are the Main Distribution and the Bayesian models, and they are implicated in the behavior encoded in the ECA.

**2.** Examples of low-level distributions are the exact moment to perform a smile, pauses between interactions, timeouts, and so forth.

The Main Distribution controls the overall behavior of the ECA. **Log Listing 1** shows the histogram of the Main Distribution over the categories after a probabilistic sampling.

## 3. Architecture

The virtual representation of KorraAI ECA has been designed using the Unity 3D framework. The current input devices are the keyboard and mouse. When a question is asked by the ECA, the user answers by either using a text box, as shown in **Figure 1**, or clicking on a button from a set of predefined answers. Support for multiple languages is provided. Verbal and nonverbal behaviors of KorraAI include speech synthesis with annotations (prosody, pauses, etc.), lip-sync, body movements, and facial expressions.

### 3.1. Initial workflow

When the ECA is started, it
- loads the ECA model (interactions, categories, distributions, and Bayesian models),
- loads all information collected from previous interactions with the user,
- sets already used interactions as "used," and
- applies "forgetfulness policy" and marks certain "used" interactions as not used.

This initial workflow is available in KorraAI's log file.

### 3.2. Creation of a new ECA model

To create a new model, the ECA designer provides the following information to KorraAI's ECA engine: More specifically, the designer needs to
- create (or reuse existing) categories of interactions,
- provide interactions for each category that optionally include reactions to user's responses,
- define a probabilistic distribution over the categories,
- define distributions over the interactions inside a category,
- choose probabilistic variables and construct Bayesian networks,
- encode changes in behavior based on the user's responses or the time that has passed using Model Triggers, and
- provide or update the distributions that govern some non-verbal interactions.

The first three points are the required minimum for a KorraAI ECA model. A programming interface forces the ECA designer to provide some of the required components. This interface also allows the ECA designer to replace and share a KorraAI ECA model with other ECA designers.

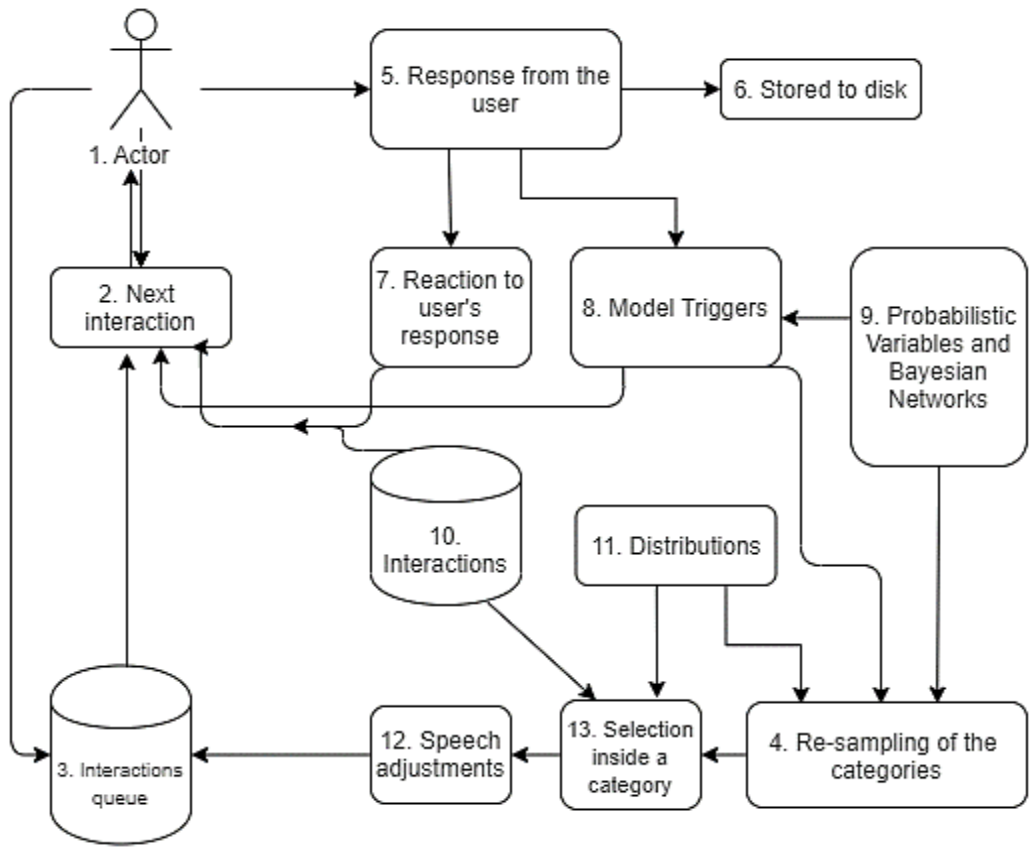

**Figure 4.** Overall view of KorraAI's architecture, with the engine on the left and the model on the right.

On the left of **Figure 4** is KorraAI's "engine", and on the right, components 7, 8, 9, 10, 11, and 12 represent the "model" provided by the ECA designer. The engine executes a loop that is defined by components 1, 2, 3, and 4. The Main Distribution is stored in component 11. The engine ensures that the interactions are loaded from 10, then uses the Main Distribution to produce the Interactions Queue 3. *Model Triggers* 8 can cause probabilistic re-sampling, as explained in Section 3.5. The *Bayesian networks* 9 are encoded using PP, and an example is given in Section 3.5.2. Bayesian networks can be used in two ways: forward and backward. In the former, we try to infer a state that is not known and depends on other nodes. In the latter, we are trying to understand the cause by knowing the result (the effect)—the probability with which a cause produced this effect. To do so, some statistics are needed to construct the probabilistic model. The ECA designer must provide them, or they can also be built during the interaction with the user. An example of backward inference is evaluating the user's level of stress (which can depend on several probabilistic nodes) without directly asking the user what his stress level is. Forward inference is used, for example, to model the number of jokes the ECA uses. Such a model is then used as part of the Main Distribution. More information is available in Appendix B. *Selection inside a category* 13 is explained in Section 3.6, and *Speech Adjustments* are described 12 in Section 3.7.

### 3.3. Base Categories

Base categories are internal structures in KorraAI that help construct the categories and sub-categories of the Main Distribution. They are not based on semantic separation but on the type of information they store. Here, we list two base categories:

- "PureFact" is used in interactions that contain exact information for either the user or the ECA. It can be used as a question or a statement. It can also contain predefined reactions depending on whether the user's response is positive or negative.

-    "UncertainFact" contains a probabilistic variable that encodes its value in the range [0,1]. It is often used in questions the ECA asks the user and usually has predefined answers that are matched to probabilities. This is explained in Section 3.4.

### 3.4. Storing a state with a level of confidence

Responses to questions with a variable degree of uncertainty are encoded with the base category UncertainFact using a probabilistic variable.

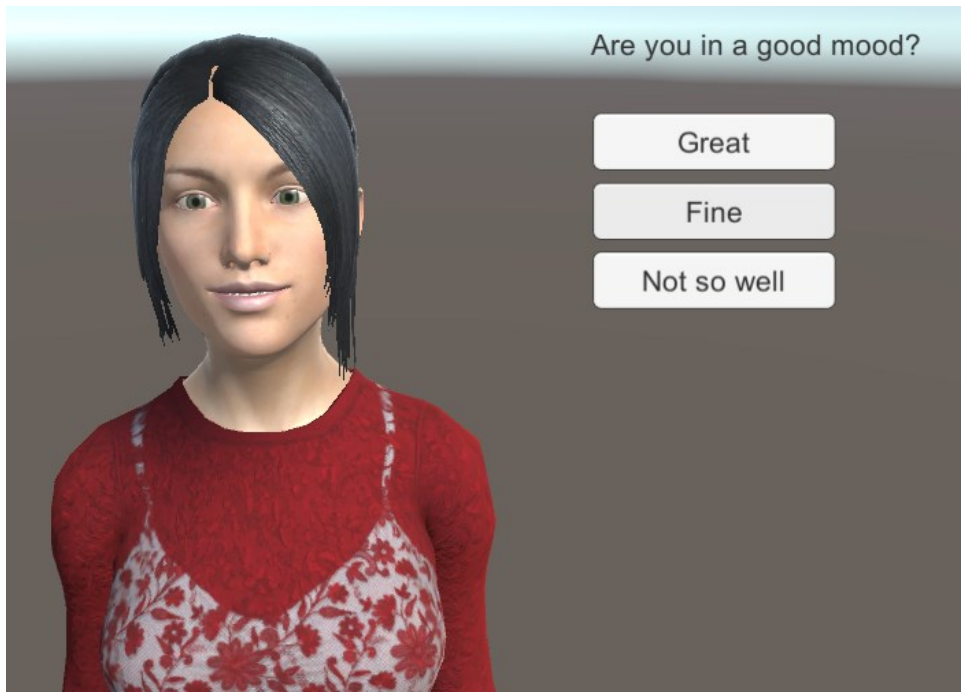

**Figure 5.** Question with three responses mapped to a probabilistic variable.

Examples of such questions are "Are you tired?" and "Are you in a good mood?" as shown in **Figure 5**. Possible responses are "Not so well," "Fine," and "Great," and they can be assigned the probabilities 0.3, 0.7, and 0.9 to describe the probabilistic variable InAGoodMood. The responses must be provided by the ECA designer along with mappings into the [0,1] range. The UncertainFact questions are usually considered repeating questions, as they usually describe a state that can change over time. These are asked periodically based on a distribution. There are two strategies in KorraAI to change the values of the probabilistic variables:
-    A probabilistic variable can assume only a limited number of values—in the case of three responses, there will be three fixed values in the range [0,1].
-    Every time the user responds, a value is added or subtracted, respecting the range [0,1].

In the second case, for example, when the user selects the response "Yes, a bit more," the probability value is increased. A negative response would reduce the probability value.

Once the answer has been mapped to probabilities, one can use, for example, InAGoodMood with a Bernoulli distribution as a part of a Bayesian network and perform inference.

### 3.5. Model Triggers

Behavior can change mainly because of two factors: (a) an external event, usually when new information from the user is available, (b) or time (time elapsed or time of the day). In KorraAI, we introduce the concept of *Model Triggers*. Two types of Model Triggers are defined (**Table 2**):

**Table 1.** Types of Model Triggers in KorraAI.

| | Can track user's responses | Can request resampling of the Main Distribution | Can add a new interaction | Can track elapsed time |
|---|---|---|---|---|
| **Model Update Trigger (MUT)** | Yes | Yes | No | Yes |
| **Model Evaluate Trigger (MET)** | Yes | No | Yes | No |

*Model Update Trigger (MUT).* MUT tracks elapsed time and the user's responses to PureFacts questions. MUT can update the Main Distribution and request a new probabilistic resampling based on it. The following is an example of modifying the rate of movie suggestions. If the user has already watched a movie, then he is probably less interested in watching a movie again. We start modeling this scenario with an interaction of type PureFact using the text "Have you watched a movie today?" which is then updated with the user's response. A predefined MUT is set to track this PureFact related to movies, and it modifies the Main Distribution in order to decrease (in the case of positive response from the user) the probability of selecting the category containing the movie suggestions. Next, the MUT will request a new resampling so that the change can take effect in future communication with the user.

*Model Evaluate Trigger (MET).* It uses a Bayesian model and the user's responses to assess a situation. The model decides whether to add a new interaction to the Interactions Queue. For instance, an example of a new interaction can be one that expresses an internal state or an emotion, such as surprise.

### 3.5.1. Changes over time using MUT

Changes in behavior over time are encoded using MUT. These changes are achieved by changing the Main Distribution and resampling it. We can have certain categories, such as Initial Communication Introduce ECA and Initial Communication Ask User, be more present in the Main Distribution at the beginning of the conversation. Then, after 10 minutes of using a MUT, we can reduce them and let other categories engage the user. Another example is a sales agent who first engages in small talk. Only after some fixed time can the agent enable another category that contains interactions designed to convince the user to buy a product.

### 3.5.2. Computing internal states of ECA using MET

As described in (44), emotions can be a strong trigger that evokes communication or a reaction from the interlocutor. Surprise is an emotion that is provoked by contradiction. It can be modeled in KorraAI using MET and a Bayesian network where the probability of observed evidence (POE) is used to calculate the credibility of the information provided by the user. The Figaro PP (45) framework provides algorithms for calculating POE. After that, a threshold can be designed based on which the system will issue a statement expressing surprise or disbelief or ask explicitly for the correct information.

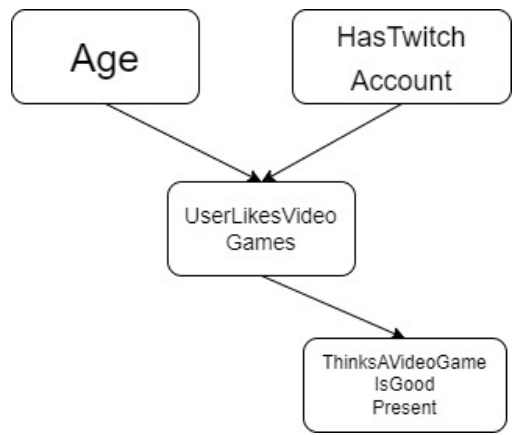

**Figure 6.** Example of encoding surprise with KorraAI.

When using a Bayesian model, an ECA designer can encode the beliefs of the ECA. The designer can set (with certain probability) that young people play video games more often, that people who have a Twitch[5] account play more games, and that playing games boosts the belief that a video game is a good present. Using the probabilistic model described in **Figure 6,** different surprise scenarios can be encoded:

- Using *Age* and *HasTwitchAccount* the model infers that *UserLikesVideoGames* has a high probability, but the user answered on the contrary when asked explicitly.

- The user answered positively to *UserLikesVideoGames* (or it was inferred) which boosts the probability of *ThinksAVideoGameIsAgoodPresent* but in the same time the user answered negatively to *ThinksAVideoGameIsAgoodPresent*.

The ECA model, Joi, implements these cases encoded in a MET in an almost identical Bayesian network. Every time new information is available from the user, the probability of contradiction is calculated. If it exceeds a threshold, a surprise interaction is inserted at the beginning of the Interactions Queue. A corresponding surprise facial expression is also triggered.

Another example of a state computation is evaluating a customer's trust in a sales agent. Such a probabilistic trust model, stored in a MET, will need to reach a certain threshold, and only then the ECA should proceed toward asking the user to buy a product by again inserting such an interaction in the Interactions Queue.

### 3.6. Interaction choice within a category

A distribution is used for sampling interactions within each category. Let us consider the joke category as an example. There can be different types of jokes: normal, romantic, dark humor, and so forth. A distribution can be used to set which jokes the ECA should start with or to explicitly reserve others for a later stage of the communication with the user. A strict rule is not imposed, so the opposite is still possible but less probable.

Another consideration is when a low number of possibilities is coupled with a low number of samples drawn (e.g., between 4 and 10) that can be used multiple times. Examples are the category Change Visual Appearance with its low number of garments or different versions of a phrase. The cognitive perception of the user should be considered. For example, repetitive sequences should be avoided. A uniform distribution with repetition can leave a single sample unused for some time. A possible strategy will be to first select a random sequence without repetition to loop over all elements and show variability and then switch to uniform sampling with repetition where the last used one is not included in the sampling.

### 3.7. Tuning of the interaction process

In our model, probabilistic sampling is used to generate most of the interactions. Next, tuning is performed in the "Speech adjustments" module. KorraAI provides an API—a set of functions to perform this tuning. Some examples of such tuning actions include

- adding an interaction at the beginning of the Interactions Queue–for example, a greeting or one provided by MET;

- grouping of interactions—for example, the introduction of the proper names of each interlocutor, exchange of courtesies; and

- removing interactions.

Instead of explicitly adding an interaction it is possible to encode with a very high probability in the Main Distribution that, for example, the first interaction should be a greeting. However, this also makes the Main Distribution heavier, and, thus, the choice between the two strategies is left to the ECA designer.

Here, we present a list of some of the techniques and strategies in KorraAI that are used to make the ECA act more naturally:

- Speech has been annotated using Speech Synthesis Markup Language (SSML). Indeed, it is extremely difficult to make a realistic ECA that employs jokes without pauses in speech, increases/decreases the speech rate, emphasizes, and so forth.

- Some dedicated distributions control whether the user is addressed by name, as in "Bob, did you know?" Another example is whether to use the phrase "OK, you know, that was a joke" (or its versions) after each joke. This clarifies intent and/or adds variability.

---

-        In the case of a limited number of interactions per category, some interactions will need to be used again. The longer time has passed, the higher the probability of reusing an interaction, as it is considered that the user does not recollect this interaction.

### 3.8. Nonverbal behaviors

One of the main features of an ECA is the ability to display nonverbal behaviors. Research on the role and benefits of using nonverbal behaviors, for example, (38-40). Several nonverbal behaviors are possible in KorraAI, and they are controlled with probabilistic distributions. An ECA designer needs to provide distributions for smiles, pauses between interactions, gaze movement, and a few others. In many cases, the normal distribution is used by default. These nonverbal behaviors are described in more detail below:

*Smiles*. One of the crucial nonverbal signals is smiling (49). According to (50), smiling agents are perceived as more likable. For example, smiles are triggered following a normal distribution with a mean of 12 seconds and a variance of 3.

*Gaze movement*. Constant staring by the ECA may have a negative impact and might not be socially accepted or even be perceived as a sign of aggressiveness (51). Thus, the agent focuses its eyes on the user at the beginning of each speech interaction for a variable amount of time after which it performs a "gaze away". This time interval follows a normal distribution with, for example, a mean of 7 seconds and a variance of 1.2. These values were selected experimentally. As pointed out in (12), an ECA should stay focused on the subject in the case of patients with psychosis, as they might interpret the gaze away as a cue of untrustworthiness.

*Pauses between the interactions*. KorraAI distinguishes between a pause before reacting to a user's response and a pause before starting a new interaction. The two types of pauses follow again the normal distribution, but with different parameters. The second distribution produces smaller time intervals as compared to the first and it has a mean of 3.7 seconds and a variance of 0.25. Waiting for a response from the user can be considered another type of pause. Again, a distribution generates the exact time before deciding that the user will not respond, and then the engine proceeds with another interaction.

The objective of these distributions is to avoid fixed time intervals.

### 3.9. Preserving the Main Distribution

The number of interactions in a category can be limited. A problem arises when a category used in the Main Distribution is depleted. The probabilities of the other categories will increase after normalization, before the probabilistic sampling. Naturally, this increase is helpful when the gap in an exhausted category needs to be filled with interactions from other categories. However, at the same time, it creates a problem when we want the usage of a certain category of interactions to be fixed in accordance with the behavior we are modeling. A solution is to mark certain categories in the model as having a fixed probability to avoid normalizing them. Only the remaining categories may change probability upon normalization.

### 3.10. Using internal statistics to build an accurate Main Distribution

KorraAI is a real-time system, and each interaction generated by the ECA takes time to complete. Usually, the Interactions Queue is filled with a fixed number of interactions following the Main Distribution. However, how much time each interaction takes needs to be considered to correctly model the Main Distribution so that the Main Distribution can be respected for a specific time interval. A KorraAI component called InteractionsStat collects and computes information about

-        the average execution time per interaction and per category;
-        the number of interactions that were requested while sampling a specific category but were unavailable due to depleted categories; and
-        forecasted interaction time (FIT)—the time the newly generated interactions would take.

The ECA must be run at least once to collect this information. To calculate the FIT, we use the following formula that loops over the categories included in the sampling $FIT = \sum_{i=1}^{n}((A_i + P_i)C_i)$, where $A_i$ is the average time spent per category; $P_i$ is the average pause time between two interactions, $C_i$ is the number of interactions for this category, and $n$ is the number of categories. $P_i$ should equal the mean parameter provided to the normal distribution that governs the pauses. The above formula is an approximation that allows us to estimate in terms of frequency and time how much a category will be present in the communication with the user and adjust the Main Distribution according to our desired model.

## 4. Evaluation and experiences

An experimental ECA application using KorraAI has been designed and published for download on the Microsoft Windows Store from October 2018 to September 2021. It should be noted that this was a feasibility study, not a controlled experiment, and we do not have a quantitative analysis.

The virtual ECA agent was called April. Anyone with a Windows 10 OS could download the application worldwide and interact with April. It had the primary objective of being an everyday companion and, secondarily, a sales agent. April uses the same categories as Joi, with an extra one called ConviceUserToBuy. Both Joi and April can ask questions, tell jokes, and provide advice and recommendations. Compared to Joi, April had a larger number of interactions in each category. April achieved 109 real sales, 3668 installations, and an average rating of 3.2/5.

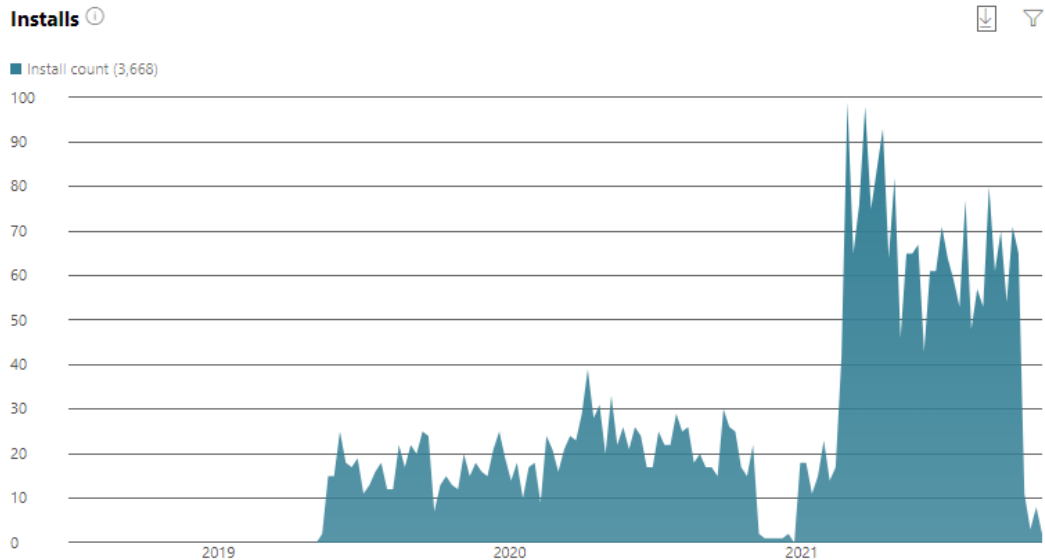

**Figure 7.** Graph of the installs. Source: screenshot from the Windows Store's menu Insights and Acquisitions.

The average installations are shown in **Figure 6**. The score improved while new functionalities were added. In our opinion, the number of installations, score, and successful sales indicate that our architecture and methods, implemented in KorraAI, performed overall satisfactorily. There are several takeaways from this study.

- ASR was one feature that was often requested by users. In KorraAI, we have intentionally left ASR out of these first versions. This is because a working ASR would stimulate the users to start asking questions freely on different topics, which could easily lead to major disappointment for the user. Moreover, an advantage in the case of a visual dialog system is that for each response the user can click, we can predefine the confidence (the probability) to which the response corresponds in a PM.

- Another aspect is that our dialog system can be improved. In addition, a request was made for the possibility of modifying the behavior. In Section 5, we address both of these.

- Another user request was for the possibility of modification of the ECA's appearance (face, hairstyle, etc.).

- There was also a complaint about the quality of the text-to-speech software (TTS). Initially, we used the TTS available in Windows 10. To improve it, we have switched to Amazon's Polly, which is of better quality and supports the Speech Synthesis Markup Language (SSML), which is largely used to make speech more natural. Amazon's Polly is a paid service.

The probabilistic reasoning performed fast enough—there were no visual glitches in the animation of the virtual actor. The system has been tested to run for hours without any crashes while using emulated responses from the user. Compared to other existing frameworks, KorraAI has relatively lower technical re-quirements, which helped with the deployment on the Windows Store. This is explained in Appendix C. During this evaluation, KorraAI was significantly improved, and using its classes and APIs, a new ECA can be easily constructed.

## 5. Discussion

Our default example ECA model, called Joi, can be used as a base model for developing new ECAs. The KorraAI framework is suitable for game designers who want diverse character modeling where the player can interact in a virtual open-world environment. KorraAI runs on the Microsoft Windows and Google Android platforms. It could potentially be used in humanoid robots and as a companion for elderly people with or without cognitive impairments.

When compared to other frameworks, KorraAI is similar to "How Was Your Day?" (HWYD) (15). Both KorraAI's Joi and HWYD have the objective of being everyday companions. While HWYD acts more like a sympathetic active listener, KorraAI always has a full Interactions Queue on how to be proactive toward the user at any moment.

KorraAI is also similar to a relational agent (RA) (12) designed for building trust and therapeutic alliances. KorraAI and this RA both use a similar UI interface where some of the user's responses are predefined on the screen. The RA used seven phases of communication, where, for example, phases 3 (assessing the patient's behavior), 4 (positive reinforcement), and 5 (tips or relevant educational materials) could be modeled in KorraAI with our concept of categories. How much each category is represented over time can be controlled by a Main Distribution. A PM of the estimated trust can be established by the ECA and Model Triggers (Section 3.5) can be used to adjust behavior of the ECA based on this model. KorraAI is also designed to be used much longer than the average ten minutes used by the RA. Indeed, the effects of the strategies employed by KorraAI are more noticeable after prolonged use of the ECA.

In KorraAI, we can have PMs for hidden states. Such a hidden state can be either one that is difficult to evaluate by the user themselves, such as stress, or one that the user does not want to admit. For example, the user might not want to admit if he is fulfilling his sports objectives for the day. Statements like "played video games all day" will indicate a significant loss of time. Such responses incorporated in a PM will indicate a deviation from the user's sports objectives. Next, a category that talks about the benefits of sports can be reinforced as part of the Main Distribution to improve the user's health state.

We are considering several future research directions and new features for KorraAI:

1) Give the user the possibility to adjust the ECA by providing a UI for the Main Distribution. This will allow the user or the ECA designer to control the ECA's behavior more easily.

2) Improve the dialog communication in KorraAI by using large language models (LLMs), which will allow the ECA to have a greater capacity for human-like communication.

3) Add ASR to KorraAI and, more specifically, intent recognition. Intent detection can be provided, for example, by the Cognitive Services Speech SDK from Microsoft or by using LLM.

4) On the probabilistic part of the framework, we plan to add another C#-compatible PPL called Microsoft Infer.NET. Infer.NET has been released as open source and provides a larger number of inference algorithms, which will be useful when using more complex PM.

5) After our preliminary experiment with KorraAI, we are currently planning a new controlled experiment to construct a sales agent or health adviser to further evaluate the effects of the probabilistic techniques employed in KorraAI.

6) To further improve the interaction with the user, we wish to include a web camera and support for virtual reality. Using image-processing techniques, we plan to detect if there is a human subject in front of the computer, count the number of persons, recognize human facial expressions, and so forth. We are also considering building the next version of KorraAI with the Babylon.js 3D framework, with the possibility of adjusting the ECA's appearance. This will allow us to avoid using closed-source components.

## 6. Conclusion

The KorraAI framework has the following contributions:

First KorraAI is an experiment on using advances in probabilistic modeling using contemporary PPLs and frameworks. A common programming language, Microsoft C#, coupled with the extension Probabilistic C#.

Other probabilistic frameworks that have been considered are Infer.NET and Figaro. Our probabilistic approach makes use of the following concepts to make the ECAs more believable:

- We employ the idea of the Main Distribution. It allows for modeling recurrent needs and desires (sleepy, tired, excited), as well as ones focused on a specific goal (e.g., get friendly with the user, sell a product). This also allows for an ECA to be executed over longer periods.

- KorraAI uses the idea of controlled indeterminism, whereby the ECA designer can set the overall behavior rather than focusing on the exact list of interactions with the user.

- Uncertain information provided by the user is naturally integrated by our architecture in the behavioral PMs.

Second, our approach focuses on making the ECA harder to get accustomed to. Even the ECA designer does not know in advance the exact interactions that will be used before the probabilistic sampling is executed based on the Main Distribution. We also use probabilistic distributions for pauses, smiles, how the speech is generated, and so on to create a complete framework with almost no fixed intervals or events. The preliminary results show that it has a rather positive effect.

Third, our architecture allows behavior to change over time, even without input from the user (e.g., based on the time of the day or elapsed time).

Forth, KorraAI is proactive, as it always has an Interactions Queue ready to be executed even without input from the user.

Fifth, we have overcome the technical difficulties of using probabilistic inference in real time, which allows for a smooth execution of the ECAs built with KorraAI. The architecture was also optimized for compatibility between components. Moreover, using our plugin architecture, ECA designers can easily exchange ECA models based on KorraAI. These are explained in Appendix C.

We hope that the architecture and experiences from our work will help ECA researchers and practitioners create more realistic ECAs.

## Appendix A. Bayesian models encoded with PPL

Several PPLs have been considered for KorraAI. Here, we present the advantages of using PPL:

- Inference and modeling are completely separate—we can define models without taking into account how we are going to make an inference.

- Probabilistic inference is built-in in the PPL.

- Models can be specified in a declarative manner.

- Models compose freely, allowing us to construct complex models from simpler ones.

- Models can be non-linear.

PPL provides a standard for encoding probabilistic models. PPL usually mixes a host language with a probabilistic extension, thus unifying general-purpose programming with probabilistic modeling. This way, we can benefit from all control structures of the host language, such as "if", "repeat", "break", and "switch", and also use probabilistic ones, such as "variable", "dist", and "flip", from the extension. As previously mentioned, we have two types of variables: normal ones that store an exact value and probabilistic ones that represent a distribution over possible values. Therefore, the probabilistic variables usually hold three components: the name of the probabilistic distribution, its parameters, and whether it is a continuous or discrete version. This is shown in **Figure 7**.

```
bool IsSprinklerOn;

Variable< bool > cloudy = Variable.Bernoulli ( 0.5 );
```

**Figure 8.** Example implementation with Infer.NET.

By using a boolean as a parameter for the class template, we specify a discrete distribution, then further specify that it will be a Bernoulli distribution, and we also set its p parameter to a value in the range [0,1].

When PPL is used, models can be easily encoded and shared. Changing the inference algorithm often requires just changing a single line of code.

KorraAI currently uses the Probabilistic C# framework (52,53). The Probabilistic C# uses monads, as explained in (52), and the LINQ functional syntax of Microsoft's C# programming language. The models presented in this article are not too complex as they do not comprise of too many probabilistic variables and evidence, so the exact inference performed by the Probabilistic C# frameworks works well. Exact inference is usually implemented as either inference by enumeration (as implemented in Probabilistic C#) or the variable elimination algorithm (VE). Exact solutions cannot scale, so bigger PMs require more advanced algorithms, which are often approximative. Probabilistic inference algorithms (both exact and approximate) can be categorized into two groups: factored algorithms, such as VE and belief propagation (BP), and sampling algorithms, such as importance sampling and the Markov chain Monte Carlo (MCMC) algorithm. Understanding these algorithms helps in designing PMs and controlling the inference process.

Modeling concepts from life with probabilities and Bayesian networks is perturbed by the following:

- One needs to decide the direction of the dependency (what causes the other), which sometimes might be difficult to evaluate.

- It is challenging to select the initial probabilities—sometimes, we need to refer to published studies; sometimes, it is a question of preference depending on the behavior we want to encode.

- Suitable inference algorithms should be used.

- Inference should be made within a reasonable amount of time.

## Appendix B. Example of forward reasoning using PPL

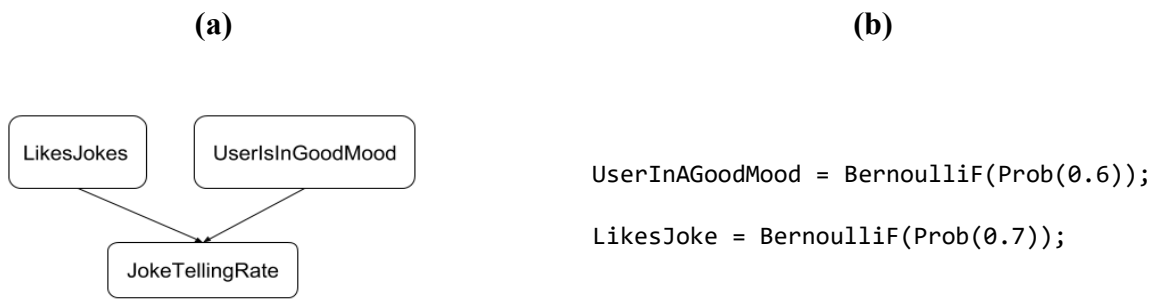

**Figure 9.** Implementation (b) of a Bayesian network describing the JokeTellingRate of an ECA (a).

When modeling a concept, we start with a graph that represents the causal relationship between variables. In **Figure 8a,** we have a Bayesian network of three probabilistic variables for the concept of how often a joke should be told by the ECA—the JokeTellingRate that is currently appropriate when communicating with the user. A Bayesian network allows for the representation of the joint probability distribution for all variables in the network through the specification of local conditional distributions corresponding to each variable. The direction of causality is usually from top to bottom. The JokeTellingRate is part of the Suggestions distributions, which is part of the Main Distribution. The code (**Figure 8b**) is in C# and describes the model.

A Bernoulli distribution is assigned to the variable UserInGoodMood. The probability of being true is 0.6, which models our a priori knowledge. The following function (**Figure 9**) describes how both LikesJokes and UserInGoodMood are used to create a truth table over the JokeTellingRate.

```
private static Func<bool, bool, Prob> TellJokeProb = (likesJoke, inGoodMood) =>

{

    if (likesJoke && inGoodMood)  return Prob(0.4)

    if (likesJoke && !inGoodMood) return Prob(0.9);

    return Prob(0.2);

};
```

**Figure 10**. A truth table over JokeTellingRate using two parameters.

If the user likes jokes and is in a good mood, we put a rather neutral value, such as 0.4. If the user likes jokes but is not in a good mood, we would like to boost the rate significantly to, for example, 0.9. The assumption here is that when one is not in a good mood, one needs more jokes to cheer up. In all the other cases, we set a low probability of 0.2. Not liking jokes means fewer jokes without considering whether the user is or is not in a good mood. The following code (**Figure 10**) defines the Bayesian network of the JokeTellingRate. It uses the truth table TellJokeProb to create a Bernoulli distribution, combined with the two probabilistic variables, LikesJoke and UserInAGoodMood, which represent our a priori knowledge.

```
public static FiniteDist<bool> JokeTellingRate =

    from like in LikesJoke

    from mood in UserInAGoodMood

    from joke in BernoulliF(TellJokeProb(like, mood))

    select joke;
```

**Figure 11**. Implementation of the JokeTellingRate. The syntax has similarities with SQL.

Next the JokeTellingRate distribution can be sampled or used as part of another Bayesian network.

## Appendix C. Details on the Technical Implementation

KorraAI is a plugin framework where the model of the ECA is compiled into a single dynamic library called KorraAI.dll. The code for the model of Joi is provided as a plugin, as shown schematically in **Figure 11**.

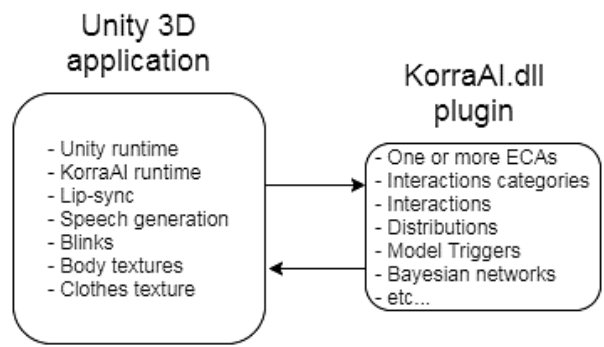

**Figure 12.** Plugin architecture.

The 3D virtual actor, together with the speech synthesis, lip-sync, music playback, and body movements, are separated from the ECA models. The separation allows the ECA designer to stay focused on the behavior and content of the ECA. The framework is compiled with the freely available Microsoft Visual Studio 2017 Community edition. Information about licensing is available in Appendix D.

KorraAI's log file is the central source of information in KorraAI. Additionally, to what is shown is **Log Listing 1** and **Log Listing 2** the KorraAI's log file contains already executed interactions, details about depleted categories and tracks the changes of probabilistic variables over time.

Testing can be challenging because the result of the system is not deterministic. At every start, a new sequence of interactions is generated using the ECA model, so we cannot directly compare it with previous sessions. We can use the log file of the ECA to verify if the generated Interactions Queue is indeed following the desired Main Distribution. Moreover, when using probabilistic inference algorithms, we can compare the results between different probabilistic frameworks to verify their correctness.

A general problem when creating an ECA framework is that machine learning, natural language processing, sentiment analysis components, and so forth, can be coded in different programming languages. Although this problem has been addressed in (9) with the GECA protocol, it is still an elaborate process. This is so because each component must be adapted to use the protocol and because, often, using such an ECA requires launching and managing several different programs in parallel, ensuring communication and synchronization between them. In addition, installing several different programs or runtimes can be a problem for a commercial ECA, especially when using a software distribution platform, such as the Microsoft Windows Store or Google Play Store, as they may not allow such a complex configuration. This is why KorraAI uses a PPL that is C#-compatible, as it is the Unity 3D framework, and thus compiles and executes as a single program. In KorraAI, we experimented using Figaro, which is a powerful PPL that comes with numerous built-in functionalities and a book (54). However, Figaro is Java-based and uses Scala as a host programming language, which dramatically increased the integration efforts and led to poor real-time performance, so it was not selected for KorraAI.

## Appendix D. Licensing

The licenses of several KorraAI components, such as the 3D avatar, lip-sync, and speech generation, prevent us from fully releasing the KorraAI framework as open source. This is why the execution engine can be provided only as a binary, which we think does not infringe on any licensing. The source code for the ECA Joi is free to use under the MIT license; it can be provided on demand, and new ECAs can be built on top of it without infringing any license.

## Acknowledgement

We are grateful to Prof. Radoslaw Niewiadomski from the University of Genoa for his thoughtful comments, suggestions and fruitful discussions.